\documentclass[10pt,twocolumn,letterpaper]{article}

\usepackage{iccv}
\usepackage{times}
\usepackage{epsfig}
\usepackage{graphicx}
\usepackage{amsmath}
\usepackage{caption}

\usepackage{times}
\usepackage{epsfig}
\usepackage{graphicx}
\usepackage{amsmath}
\usepackage{amssymb}
\usepackage{float}
\usepackage{caption}
\usepackage{subcaption}
\usepackage{booktabs} 
\usepackage{textcomp}
\usepackage{multirow}
\usepackage{amssymb}
\usepackage{makecell}
\usepackage{notoccite}
\usepackage{comment}
\usepackage{subfiles}

\def \nAnime {1,972}
\def \nFrame {122,365}
\def \nCategory {31}

\usepackage[dvipsnames]{xcolor}
\definecolor{mygray}{gray}{0.6}
\newcommand{\blue}[1]{{\color{blue}#1}}
\newcommand{\red}[1]{{\color{red}#1}}

\usepackage[pagebackref=true,breaklinks=true,letterpaper=true,colorlinks,bookmarks=false]{hyperref}

\iccvfinalcopy

\ificcvfinal\fi

\begin{document}

%%%%%%%%% TITLE
\title{4DComplete: Non-Rigid Motion Estimation Beyond the Observable Surface}

\author{
Yang Li${}^\text{1} $\quad 
Hikari Takehara${}^\text{2}$\quad 
Takafumi Taketomi${}^\text{2}$\quad  
Bo Zheng${}^\text{2}$\quad 
Matthias Nie{\ss}ner${}^\text{3}$ \vspace{0.2cm} \\
${}^\text{1}$The University of Tokyo\quad  
${}^\text{2}$Tokyo Research Center, Huawei\quad  
${}^\text{3}$Technical University Munich \vspace{0.2cm} \\
}

\newcommand\blfootnote[1]{%
  \begingroup
  \renewcommand\thefootnote{}\footnote{#1}%
  \addtocounter{footnote}{-1}%
  \endgroup
}

\twocolumn[{%
	\renewcommand\twocolumn[1][]{#1}%
	\maketitle
	\begin{center}
		\vspace{-0.8cm}
		\captionsetup{type=figure}
		\includegraphics[width=0.96\linewidth]{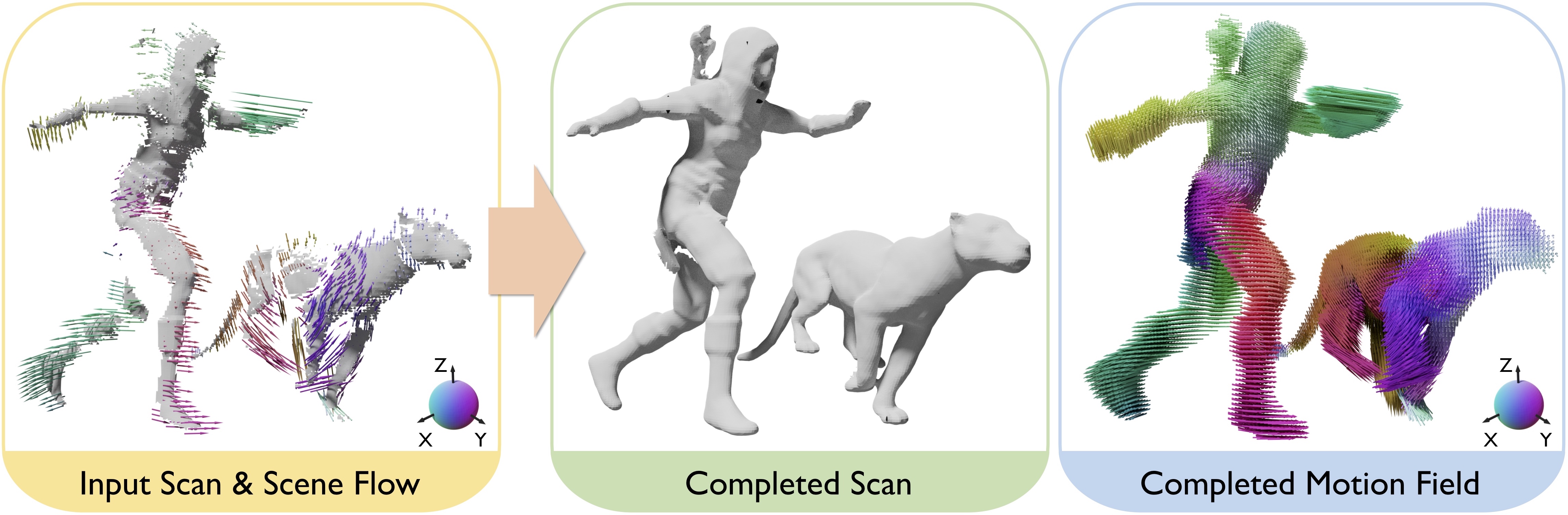}
		\vspace{-0.2cm}
		\captionof{figure}{
Given an input partial scan and inter-frame scene flow from a non-rigidly deforming scene (left),
our method jointly recovers missing geometry (middle) and a volumetric motion field (right). 
The colors show the motion directions on the unit sphere (bottom right corner) where the vector's length corresponds to the magnitude of the motion.
}
		\label{fig:teaser}
	\end{center}  
}]

% Remove page # from the first page of camera-ready.
\ificcvfinal\thispagestyle{empty}\fi

\begin{abstract}
Tracking non-rigidly deforming scenes using range sensors has numerous applications including computer vision, AR/VR, and robotics.
However, due to occlusions and physical limitations of range sensors, existing methods only handle the visible surface, thus causing discontinuities and incompleteness in the motion field.
To this end, we introduce 4DComplete, a novel data-driven approach that estimates the non-rigid motion for the unobserved geometry.
4DComplete takes as input a partial shape and motion observation, extracts 4D time-space embedding, and jointly infers the missing geometry and motion field using a sparse fully-convolutional network.
For network training, we constructed a large-scale synthetic dataset called DeformingThings4D, which consists of \nAnime~animation sequences spanning \nCategory~different animals or humanoid categories with dense 4D annotation. Experiments show that 4DComplete 1) reconstructs high-resolution volumetric shape and motion field from a partial observation, 2) learns an entangled 4D feature representation that benefits both shape and motion estimation, 3) yields more accurate and natural deformation than classic non-rigid priors such as As-Rigid-As-Possible (ARAP) deformation, and 4) generalizes well to unseen objects in real-world sequences.
\end{abstract}

\section{Introduction}
%Problem Statement / Motivation:
Understanding the motion of non-rigidly deforming scenes using a single range sensor lies at the core of many computer vision, AR/VR, and robotics applications. 
In this context, one fundamental limitation is that a single-view range sensor cannot capture data in occluded regions, leading to incomplete observations of a 3D environment.
As a result, existing non-rigid motion tracking methods are restricted to the observable part of the scene.
However, the ability to infer complete motion from a partial observation is indispensable for many high-level tasks.  
For instance, as a nursing robot, to safely care for an elderly person (e.g., predict the person’s action and react accordingly), it needs to understand both the complete body shape and how the whole body moves even if the person is always partially occluded.

In order to address these challenges, we pose the question
\textit{how can we infer the motion of the unobserved geometry in a non-rigidly deforming scene?} 
Existing works such as DynamicFusion~\cite{dynamicfusion} and VolumeDeform~\cite{volumedeform} propose to propagate deformations from the visible surface to the invisible space through a latent deformation graph.
Hidden deformations are then determined by optimizing hand-crafted deformation priors such as As-Rigid-As-Possible~\cite{arap} or Embedded Deformation~\cite{embededdeformation}, which enforces that graph vertices locally move in an approximately rigid manner.
Such deformation priors have several limitations: 1) they require heavy parameter tuning; 2) they do not always reflect natural deformations; and 3) they often assume a continuous surface. 
As a result, these priors are mostly used as regularizers for local deformations, but struggle with larger hidden regions.
One promising avenue towards solving this problem is to leverage data-driven priors that learn to infer the missing geometry. 
Very recently, deep learning approaches for 3D shape or scene completion and other generative tasks involving a single depth image or room-scale scans have shown promising results~\cite{3d_epn,SSCNet_shuran,scancomplete,chibane20ifnet,sgnn_cvpr2020}. 
However, these works primarily focus on static environments.

%Our Method: what we do
In this paper, we make the first effort to combine geometry completion with non-rigid motion tracking. 
We argue that the shape and motion of non-rigidly deforming objects are highly entangled data modalities:
on one hand, the ability to infer the geometry of unobserved object parts provides valuable information for motion estimation.
On the other hand, motion is considered as the shape's evolution in the time axis, as similarity in motion patterns are a strong indicator for structural connectivity. 
To leverage these synergies, we propose 4DComplete, which jointly recovers the missing geometry and predicts motion for both seen and unseen regions. 
We build 4DComplete on a sparse, fully-convolutional neural network, which facilitates the joint estimation of shape and motion at high resolutions.
In addition, we introduce DeformingThings4D, a new large-scale synthetic dataset which captures a variety of non-rigidly deforming objects including humanoids and animals. 
Our dataset provides holistic 4D ground truth with color, optical/scene flow, depth, signed distance representations, and volumetric motion fields.

\medskip
\noindent
In summary, we propose the following contributions:
\begin{itemize}
\item 
We introduce 4DComplete, the first method that jointly recovers the shape and motion field from partial observations.
\item
We demonstrate that these two tasks help each other, resulting in strong 4D feature representations outperforming existing baselines by a significant margin.
\item 
We provide a large-scale non-rigid 4D dataset for training and benchmarking.
The dataset consists of \nAnime~animation sequences, and \nFrame~frames.
Dataset is available at: \url{https://github.com/rabbityl/DeformingThings4D}.
\end{itemize}

\section{Related Work}

\subsection{Non-Rigid Tracking Using Depth Sensors}

\begin{figure*}[!ht]
\centering
\includegraphics[width= 1\linewidth  ]{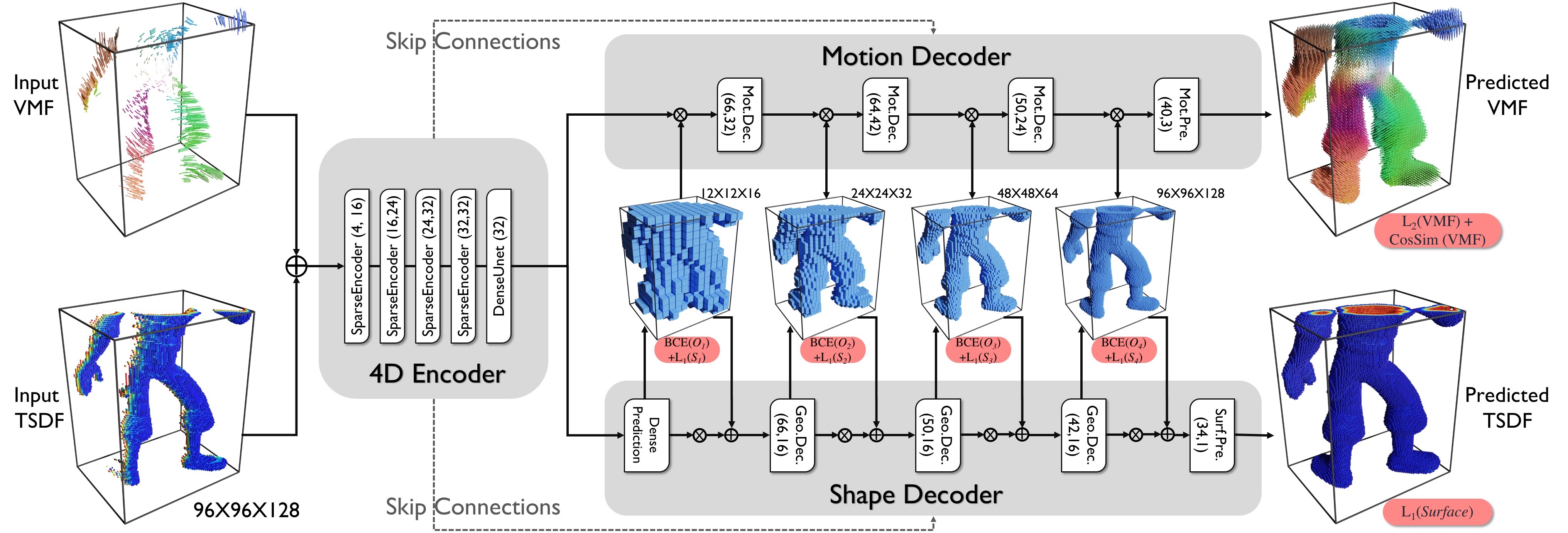}
\caption{
The network architecture of 4DComplete (Pink capsule: training loss; $\mathbf{\oplus}$: concatenation; $\mathbf{\otimes}$: filter by geometry, number in the brackets: ($n_{in}$, $n_{out}$) feature dimension).
The input partial TSDF and VMF are concatenated together and fed into the 4D encoder. The two decoders predict the complete TSDF and VMF in parallel.  
There are 4 hierarchical levels. The shape decoder predicts the geometry in each hierarchical level and passes the predicted geometry to the corresponding layer in the motion decoder and the following layer in the shape decoder.
Our method is trained on cropped volumes of spatial dimension $96\times96\times128$, which covers about 70 percent of an object. The fully-convolutional nature of our approach enables testing on whole objects of arbitrary sizes.
}
\label{fig:network_architecture}
\end{figure*}

Many methods for non-rigid tracking use variations of the Non-rigid Iterative Closest Point (N-ICP) algorithm~\cite{NICP-1,NICP-2,li2008regist,zollhofer2014real}, where the point-to-point or point-to-plane distance of correspondences points are iteratively minimized.
To prevent uncontrolled deformations and resolve motion ambiguities, the N-ICP optimization usually employs deformation regularizers such as As-Rigid-As-Possible (ARAP)~\cite{arap} or embedded deformation~\cite{embededdeformation}.
One of the first real-time methods to jointly track and reconstruct non-rigid surfaces was DynamicFusion~\cite{dynamicfusion}.
VolumeDeform~\cite{volumedeform} extends the ideas of DynamicFusion by adding sparse SIFT feature matches to improve tracking robustness.
Using deep learning, DeepDeform~\cite{deepdeform} replaces the classical feature matching by CNN-based correspondence matching.
Li et al.~\cite{Learning2Optimize} goes one step further and differentiates through the N-ICP algorithm thus obtaining a dense feature matching term.
A similar direction is taken by Neural Non-Rigid Tracking~\cite{bovzivc2020neural}; however, their focus lies in an end-to-end robust correspondence estimation.
To handle topology changes, KillingFusion~\cite{killingfusion} directly estimates the motion field given a pair of signed distance fields (SDF).
Optical/scene flow~\cite{flownet,RAFT_eccv2020,pwcnet,global_patch_collider,flownet3d,wu2019pointpwc,liu2019meteornet,HPLFlownet,lv2018LearningRigidity} is a closely related technique.
They have been used to generate initial guess for non-rigid tracking in~\cite{dou2016fusion4d,surfelwarp,bovzivc2020neural,motion2fusion,flownet3d++}. 
Among these works, FlowNet3D~\cite{flownet3d} is one of the first methods that directly estimates scene flow from two sets of point clouds.  
While existing methods mainly focus on the visible surface of a scene, we take one step further to model the deformation of the hidden surface. 

\subsection{Shape and Scene completion}

Completing partial 3D scans is an active research area in geometry processing. 
Traditional methods, such as Poisson Surface Reconstruction~\cite{PoissonSurfaceReconst}, locally optimize for a surface to fit observed points and work well for small missing regions.
Zheng et al.~\cite{zheng2013beyond} predict the unobserved voxels by reasoning physics and Halimi et al.~\cite{WholeIsGreaterThanParts} complete partial human scans by deforming human templates.
More recently, we have seen 3D CNNs with promising results for geometry completion for depth scans ~\cite{SSCNet_shuran,3d_epn,scancomplete,sgnn_cvpr2020}. 
These works operate either on a single depth image of a scene as with SSCNet~\cite{SSCNet_shuran}, or scene completion on room- and building floor-scale scans, as shown by  ScanComplete~\cite{scancomplete} and SGNN~\cite{sgnn_cvpr2020}.
An alternative line of research for shape completion uses implicit scene representations~\cite{occupancy_network,occupancy_flow,convolutional_occnet,deepsdf,PiFusion,chibane20ifnet,chiyu2020localimplicit}; however, while these approaches achieve stunning results for fitting and interpolation of objects/scenes, they still struggle to generalize across object categories with high geometric variety.
While these existing works mainly focus on static scenes, we investigate how to leverage the synergies of shape completion in the dynamic 4D domain.

\subsection{Non-Rigid 4D Datasets}

Collecting large scale 4D datasets for deforming objects is a non-trivial task, in particular when the goal is to obtain a sufficiently large number of objects.
Non-rigid datasets~\cite{de2008performance,faust,anguelov2005scape,vlasic2008articulated,guo2015robust,ye2012performance,volumedeform,killingfusion,deepdeform,Zheng2019DeepHuman,AMASS_ICCV2019,3D_menagerie} have been widely used, but they are either relatively small, limited to specific scene types, or suffer from occlusion and sensor noise; hence, they are not directly suited for our 4D completion task.
Notably, obtaining dense motion field ground truth from real-world 3D scans is quite challenging as it requires costly per-point correspondence annotations. 
This is one of the reasons why we have seen many synthetic datasets in the context of dense optical flow methods ~\cite{Monka,Sintel,lv2018LearningRigidity,playing_for_benchmarks};
Among them, Sintel~\cite{Sintel} and Monka~\cite{Monka} are composed of rendered animations of deforming objects. 
However, these sequences are relatively short and do not provide complete 3D shapes and motion fields.  
In order to facilitate learning data-driven deformation priors, we introduce a much larger synthetic dataset with over \nAnime~animation sequences, spanning a large diversity of objects ranging from humanoids to various animal species (cf. Sec.~\ref{section-dataset}).

\section{Method: 4DComplete}
Given a single-view depth map observation of a 3D scene, and the scene flow that is computed between the current frame and its next frame, the goal of 4DComplete is to recover the hidden geometry and its motion field.

\bigskip
\noindent
\textbf{Input.} 
We use a 3D volumetric grid to represent both shape and motion. 
The input shape is represented as a truncated signed distance field (TSDF), as a sparse set of voxel locations within truncation and their corresponding distance values. 
The TSDF is computed from a single depth map using  volumetric fusion~\cite{volumetric_fusion}; i.e., every voxel is projected into the current depth map and their distance values are updated accordingly.
To represent the input motion of the visible surface, we pre-compute the 3D motion vector (in $\mathbb{R}^3$) for each occupied voxel, resulting in a volumetric motion field (VMF) representation.
We concatenate the TSDF and VMF and feed it as input to a neural network.

\bigskip
\noindent
\textbf{Scene Flow Field $\Leftrightarrow$ Volumetric Motion Field.} 
We use FlowNet3D~\cite{flownet3d} to predict the motion for the visible surface, which estimates the scene flow field (SFF) between two sets of point clouds. 
Because a 3D point does not necessarily lie on a regular 3D grid position, we convert between the SFF and the VMF as follows:
given a point cloud $\{  {p_i|i = 1,...,N} \} $, where $p_i \in \mathbb{R}^3$ are XYZ coordinates of individual point, the SFF is defined as $ \{ \mathcal{SFF}_i|i = 1,...,N \} $, where $\mathcal{SFF}_i\in \mathbb{R}^3$ are the 3D translational motion vectors of the points. 
Similarly, given a set 3D voxel positions  $\{ {v_j|j = 1,...,M} \} $, the VMF is defined as $ \{ \mathcal{VMF}_i|i = 1,...,M \} $, where $\mathcal{VMF}_i\in \mathbb{R}^3$ are the 3D translational motion vectors of the voxels. To convert from SFF to VMF, we use the inverse-distance weighted interpolation as defined in \cite{pointnet++}: 
\begin{equation}
\label{eqn:sff_2_vmf}
\mathcal{VMF}_j =  \sum_{p_i \in knn ( v_j )  }  \frac{ \mathcal{SFF}_i \cdot dist(p_i, v_j)^{-1} }{\sum_{p_{i} \in knn ( v_j )  }dist(p_i, v_j)^{-1}}
\end{equation}  
where $knn()$ is the function to find K-Nearest-Neighbors.
We set the number of neighbors to $K=3$, and $dist(,)$ computes the euclidean distance between two positions. 
To convert from VMF to SFF, we do tri-linear interpolation:
\begin{equation}
\label{eqn:vmf_2_sff}
\mathcal{SFF}_j =  \sum_{v_j \in knn ( p_i )  }  \mathcal{VMF}_j \cdot w(p_i, v_j)
\end{equation} 
where $w(,)$ computes the linear-interpolation weights, and $K=8$ represents the neighboring 8 corners of the cube that a point lies in.

\bigskip
\noindent
\textbf{Network Architecture.}
To allow for high-resolution outputs of the shape and motion field, we leverage sparse convolutions~\cite{sparseconv_arxiv,sparseconv_iccv,choy2019minkowski} for our neural network architecture, which makes our architecture computationally efficient in processing 3D volumetric data by operating only on the surface geometry.
Hence, our method only processes the surface region and ignores the truncated region.
Fig.~\ref{fig:network_architecture} shows an overview of our network architecture. 
The network consists of a shared 4D encoder and two decoders to estimate shape and motion in parallel. 
The input sparse tensor is first fed into the~\textit{4D Encoder}, which encodes the data using a series of sparse convolutions where each set reduces the spatial dimensions by a factor of two.
The two decoders are designed in a coarse-to-fine architecture with 4 hierarchical levels.
We use skip connections between the 4D encoder and the 2 decoders to connect feature maps of the same spatial resolution. 
Since the shape decoder usually generates a larger set of sparse locations than the input, we use zero feature vector for the locations that do not exist in the input volume.

\bigskip
\noindent
\textbf{Message passing between two branches.}
At a hierarchical level $k$, the shape decoder predicts the voxels' occupancy $O_k$ and TSDF value $S_k$.  
We filter voxels with $sigmoid(O_k (v)) > 0.5$ as the input geometry for the next hierarchical level.
Within each hierarchical level, the shape decoder feeds the predicted geometry to the parallel motion decoder to inform where the motion should be estimated. In return, the motion feature is filtered by the sparse geometry and shared to the shape decoder.

\bigskip
\noindent
\textbf{Shape Loss.}
The shape decoder's final output is a sparse TSDF from which a mesh can be extracted by Marching Cubes.
Following~\cite{sgnn_cvpr2020}, we apply an $l_1$ loss on the log-transformed TSDF values. 
Using the log-transformation on the TSDF values helps to shift the losses attention more towards the surface points as larger values far away from the surface get smaller, thus encouraging more accurate prediction near the surface geometry.
We additionally employ proxy losses at each hierarchy level for outputs $O_k$ and $S_k$, using binary cross-entropy with target occupancies and $l_1$ with target TSDF values, respectively.

\bigskip
\noindent
\textbf{Motion Loss.}
The output of our sparse neural network is facilitated by the motion decoder, which estimates the completed volumetric motion field $ \{ \mathcal{VMF}_i|i = 1,...,M \} $.
The ground truth motion field at the predicted sparse locations is given by $ \{ \mathcal{VMF}_{i,gt}|i = 1,...,M \} $. 
We formulate the loss for the motion field on the final predicted sparse locations using the $l_2$ loss:  $\sum_{i=1}^M ||\mathcal{VMF}_i - \mathcal{VMF}_{i,gt} ||_2^2$. 
In addition, we apply the cosine similarity loss: 
$\sum_{i=1}^M (1 -  \frac{\mathcal{VMF}_i \cdot \mathcal{VMF}_{i,gt} }{ ||\mathcal{VMF}_i ||\cdot ||\mathcal{VMF}_{i,gt}||})$
on the normalized motion vectors to encourage the directions of the motion to be consistent with the ground truth.

\bigskip
\noindent
\textbf{Progressive Growing.}
We train our network in a progressively growing fashion following the ideas of \cite{sgnn_cvpr2020}.
There are four hierarchy levels, we progressively introduce higher resolution geometry decoder after every 2000 training iterations.
To facilitate motion decoder learning, instead of using the predicted geometry of shape decoder, we fed ground-truth geometry to motion decoder during the beginning 10K iterations.

\bigskip
\noindent
\textbf{Training.}
We use our newly-constructed DeformingThings4D dataset (c.f. Sec.~\ref{section-dataset}) to train our network.
At training time, we consider cropped views of scans for efficiency (see Fig.~\ref{fig:network_architecture}); we use random crops of size $[96\times 96\times 128]$ voxels for the finest level. 
We crop the volumes at $1$ meter intervals out of each train object and discard empty volume.
The resolution drops by a factor of 2, resulting resolution of $[48\times 48\times 64]$, $[24\times 24\times 32]$, and $[12\times 12\times 16]$ for each hierarchical level.
The fully-convolutional design of our approach enables testing on whole objects of arbitrary sizes at testing time.
To learn viewpoint-invariant motion representation, we apply random rigid rotation transformations on the 3D motion vectors as data augmentation during training. 
The randomness is drawn from the Haar distribution~\cite{random_orthogonal}, which yields uniform distribution on SO3.
We train our network using the Adam optimizer with a learning rate of 0.001 and a batch size of 8.

\section{DeformingThings4D Dataset} \label{section-dataset}
\begin{figure*}[!t]
\centering
\hspace*{-0.7cm} 
\includegraphics[width= 1.05\linewidth]{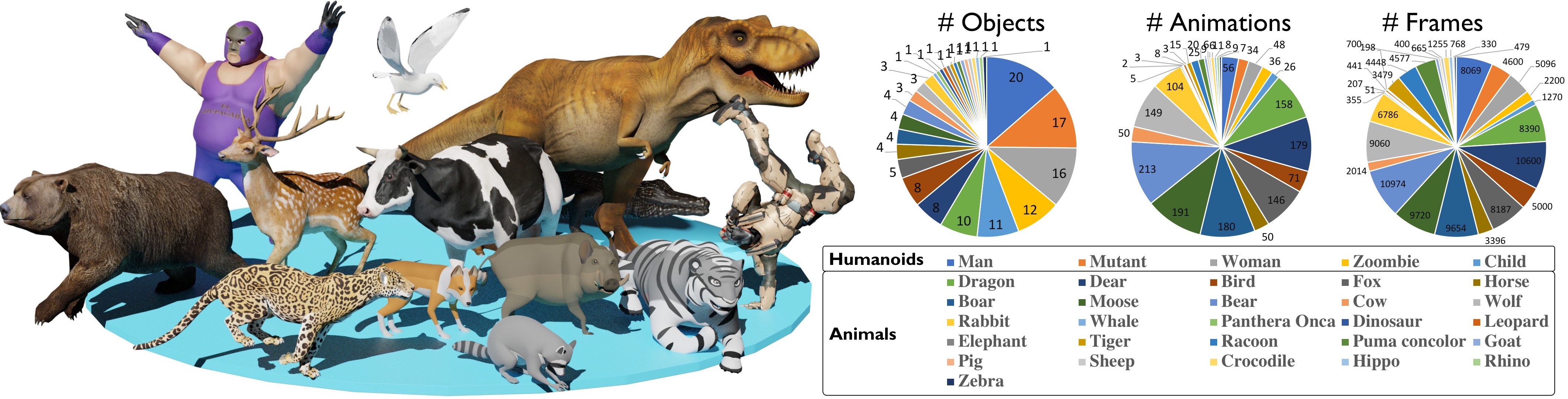}
\caption{
DeformingThings4D dataset. 
Left: examples of animated characters. 
Right: dataset statistics. In total, we collected 147 different characters spanning \nCategory~categories, with a total of \nAnime~animations
and \nFrame~frames.
}
\label{fig:Dataset_examples}
\end{figure*}

Training our network requires a sufficient amount of non-rigidly deforming target sequences with ground truth 4D correspondences at the voxel level (i.e., motion and shape).
In order to provide such data, we construct a synthetic non-rigid dataset, DeformingThings4D, which consists of a large number of animated characters including humanoids and animals with skin mesh, texture, and skeleton.
We obtained the characters from Adobe Mixamo\footnote{https://mixamo.com} where humanoid motion data was collected using a motion capture system. 
Animals' skin and motion are designed by CG experts.  
Generally, these objects are animated by using ``rigging" and ``skinning" to blend the skeletal movement to the surface skin mesh. 
Fig.~\ref{fig:Dataset_examples} shows examples of characters in the dataset and the statistics of our dataset.

\subsection{Data Generation}\label{section:data_generation}
\begin{figure}[!h]
\centering
\includegraphics[width= 1\linewidth]{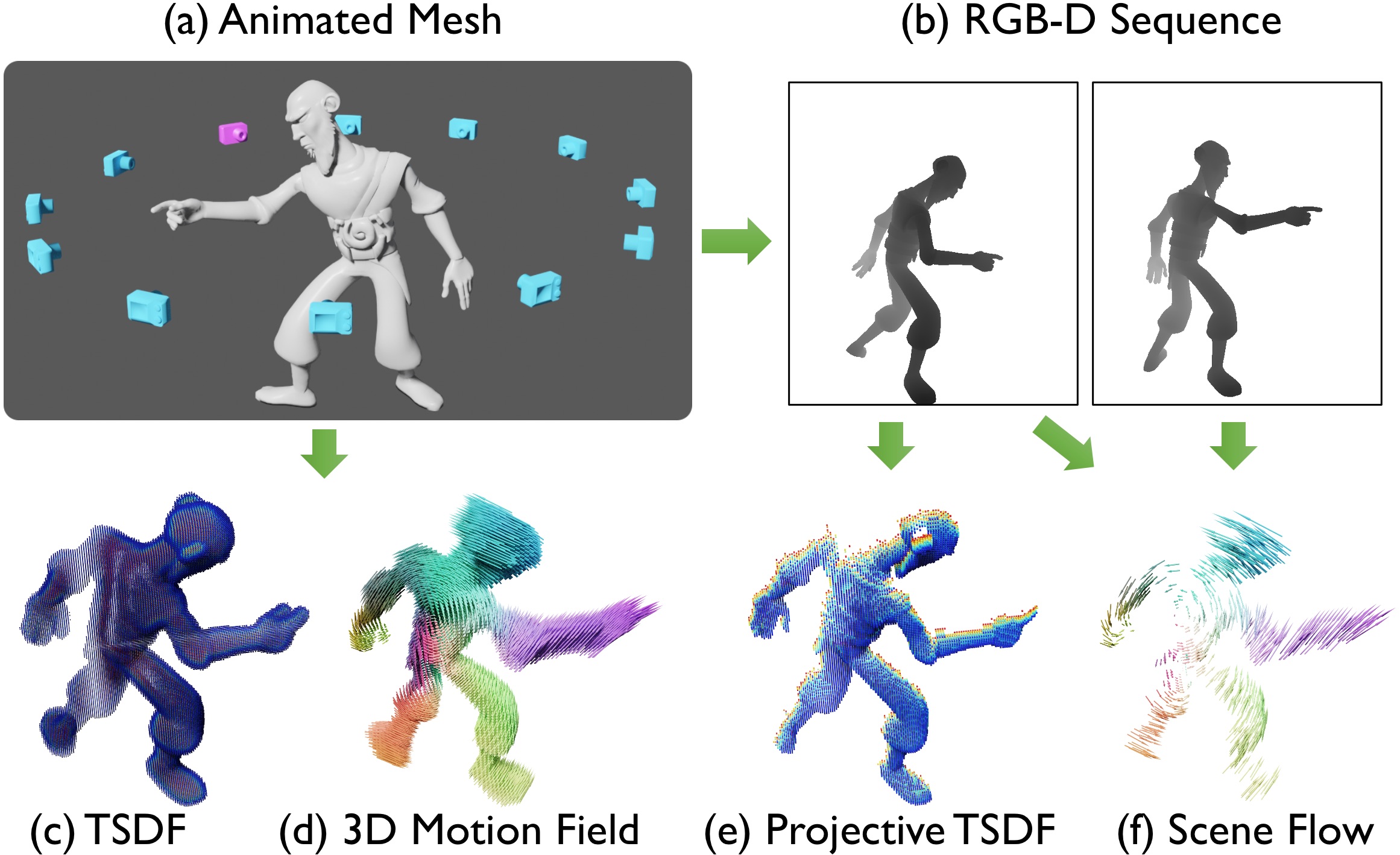}
\caption{
Data Generation Process: 
given an animated 3D mesh (a), virtual cameras are sampled on a sphere. One of the cameras is selected as the input view, for which depth maps (b) are rendered. 
Depth frames are used to compute the projective TSDF (e) and inter-frame scene flow (f).
The ground-truth complete TSDF (c) is computed by integrating the depth images from all virtual cameras. The complete 3D motion field (d) is obtained by blending the mesh vertices' motion to nearby occupied voxels. 
}
\label{fig:datagen}
\end{figure}

Given a animated 3D mesh, we generate per-frame RGB-D maps, inter-frame scene flow, signed distance field and volumetric motion field; see Fig.~\ref{fig:datagen}.
We perform data generation with Blender\footnote{https://www.blender.org/} scripts.

\medskip
\noindent
\textbf{RGB-D Map.} To render depth maps, we uniformly sample 42 camera viewpoints on a sphere that is centered by the target character's mesh.
The mesh-to-camera distance ranges within $0.5-2.5m$. 
We render all the depth maps using the intrinsic parameters of the Azure Kinect camera. 
We store per-pixel depth in millimeters and render the color channel using Blender's built-in Eevee engine with a principled BSDF shader.

\medskip
\noindent
\textbf{Inter-frame Scene Flow Field (SFF).} 
The mesh animations run at 25 frames per second. We track the mesh vertices' 3D displacements between a pair of temporally adjacent frames and project the 3D displacements to the camera's pixel coordinates as scene flow.  
The flow vector for a pixel is computed by interpolating the 3 vertices' motion on a triangle face where the pixel's casted ray is first received.
We generate scene flow ground truth for all observable pixels in the source frame even if the pixels are occluded in the target frame. To simulate the different magnitudes of deformation we sub-sample the sequences using the frame jumps: \{1, 3, 7, 12\}.

\medskip
\noindent
\textbf{Signed Distance Field (SDF).} 
In order to generate the ground truth SDF, we volumetrically fuse the depth maps from all virtual cameras into a dense regular grid~\cite{volumetric_fusion}, where each voxel stores a truncated signed distance value.  
We repeat this process independently for four hierarchy levels, with voxel sizes of  $1.0cm^3$, $2.0cm^3$, $4.0cm^3$, and $8.0cm^3$.
From the input depth map, we compute the projective SDF with voxel sizes of $1.0cm^3$ as network input while setting the truncation to 3$\times$ the voxel size. 
TSDF values are stored in voxel-distance metrics, which facilitates testing on volumes with arbitrarily sampled voxel size.

\medskip
\noindent
\textbf{Volumetric Motion Field (VMF).} 
We compute the motion ground-truth for all voxels near the mesh surface, i.e., within 3x voxel truncation. 
For each valid voxel, we first find its $K$-nearest-neighbor vertices on the mesh surface and then use Dual Quaternion Blending (DQB) to bind the motion of the KNN vertices to the voxel position. Empirically, we set $K=3$.
We follow the same procedure for the SDF volume and we repeat this process independently for all four resolutions, i.e., with voxel size of $1.0cm^3$, $2.0cm^3$, $4.0cm^3$, and $8.0cm^3$.

\begin{table*}[!ht]
\centering
\small
\renewcommand{\arraystretch}{1} % Default value: 1
\setlength{\tabcolsep}{3pt} % Default value: 6pt

\begin{tabular}{|l|c|c|c|c|c|c|c|c|c|c|} \hline   
       && \multicolumn{3}{c|}{DeformingThings4D} & \multicolumn{3}{c|}{DeepDeform Dataset~\cite{deepdeform}}  & \multicolumn{3}{c|}{KITTI SFLow~\cite{kitti}}\\ 
\cline{3-11} 

Method& Training dataset   & EPE$\downarrow$      & ACC(5)$\uparrow$     & ACC (10)$\uparrow$    & EPE$\downarrow$   & ACC(5)$\uparrow$  & ACC(10)$\uparrow$  &EPE$\downarrow$   & ACC(5)$\uparrow$  & ACC(10)$\uparrow$ \\ \hline
 
\multirow{ 2}{*}{FlowNet3D}& FlyingThings3D~\cite{Monka}   & 7.36  &  69.43\%     &  80.04\%       &  21.07  &  24.62 \%    & 45.09\%        &\textbf{16.88} &\textbf{38.49\%} &67.17\%   \\  \cline{2-11} 
 & DeformingThings4D (Ours) &\textbf{3.74}     & \textbf{82.02\% }     & \textbf{91.63}\%       &  \textbf{13.08}  &  \textbf{27.78 \%}    & \textbf{61.26\%}     &17.01 &36.89\% &\textbf{71.67}\%      \\ \hline

\end{tabular}
\caption{
Scene flow estimation results on the DeformingThings4D, DeepDeform~\cite{deepdeform}, and KITTI~\cite{kitti} datasets.
Metrics are end-point-error (EPE) in centimeters, and Accuracy ( \textless 5$cm$ or 5\%, 10$cm$ or 10\%) for motion.
}
\label{tab:dataset_compare}
\end{table*}

\begin{figure}[!h]
\centering
\includegraphics[width= 1\linewidth ]{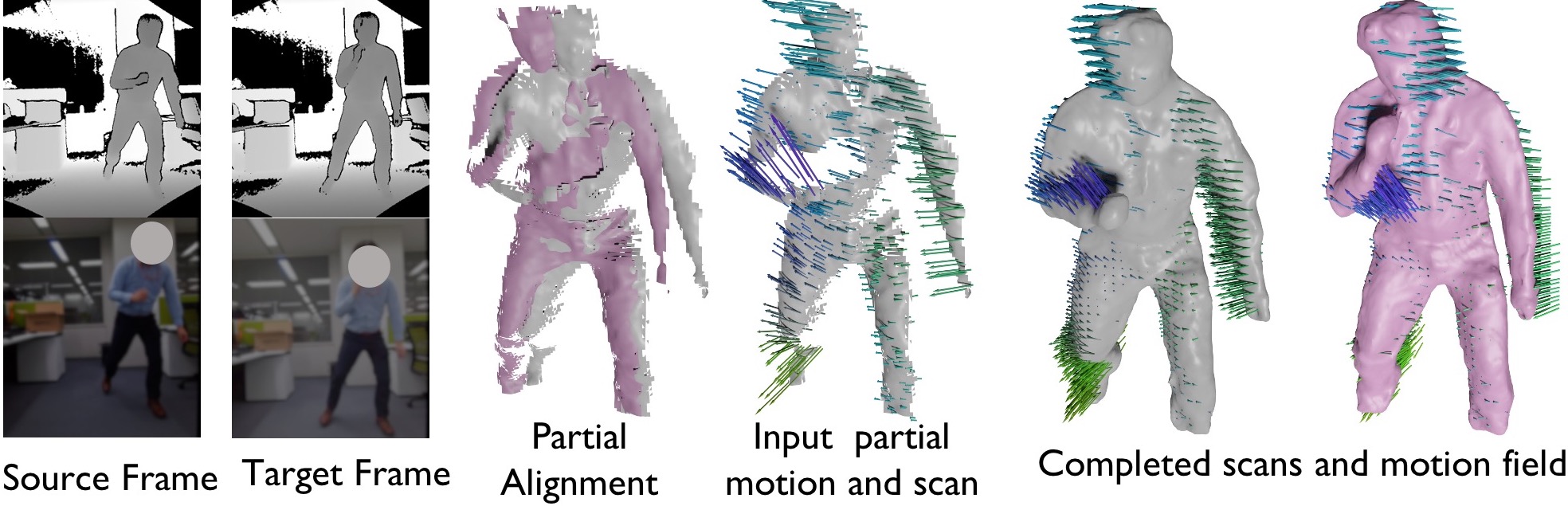}
\caption{
Testing on a pair of real-world RGB-D images. 
}
\label{fig:demo_screenshot}
\end{figure}

\section{Results}

\subsection{Evaluation Metrics}

\smallskip
\noindent
\textbf{Motion Estimation Evaluation Metric.} 
Following~\cite{flownet3d}, we use 3D end-point-error (EPE) and motion accuracy (ACC) as our motion evaluation metrics. 
The 3D EPE measures the average euclidean distance between the estimated motion vector to the ground truth motion vector. 
The ACC score measures the portion of estimated motion vectors that are below a specified end-point-error among all the points. 
We report two ACC metrics with two different thresholds.
Note that throughout the experiments we convert all VMF to SFF (using Eqn.~\ref{eqn:vmf_2_sff}) before doing motion evaluation. 

\medskip
\noindent
\textbf{Shape Completion Evaluation Metric.} 
We use the following metrics to evaluate the reconstructed geometry, Volumetric IoU (IoU), Chamfer Distance (CD) in centimeters, Surface Normal Consistency (SNC), Point to Plane distance (P2P), and $\ell_1$ error of SDF value.

\subsection{Benchmarking Scene Flow} 
We compare our DeformingThings4D dataset with the FlyingThings3D~\cite{Monka}, which is a large-scale dynamic motion dataset consists of flying rigid objects.
We train FlowNet3D~\cite{flownet3d} with the two datasets and evaluate it on the test sets of DeformingThings4D, the DeepDeform~\cite{deepdeform}, and the KITTI~\cite{kitti} scene flow benchmark.
The results are shown in Tab.~\ref{tab:dataset_compare}.
DeepDeform~\cite{deepdeform} is a very challenging real-world benchmark for non-rigid motion.
The FlowNet3D model trained on our dataset significantly reduces the scene flow error on the real-world DeepDeform benchmark (from 21.07 to 13.08). 
KITTI dataset captures street scenes with mainly rigid cars moving around which is more close to the flying things scenario.
Our dataset still shows comparable results to FlyingThings3D on KITTI.

\begin{figure*}[!h]
\centering
\includegraphics[width= 0.89\linewidth ]{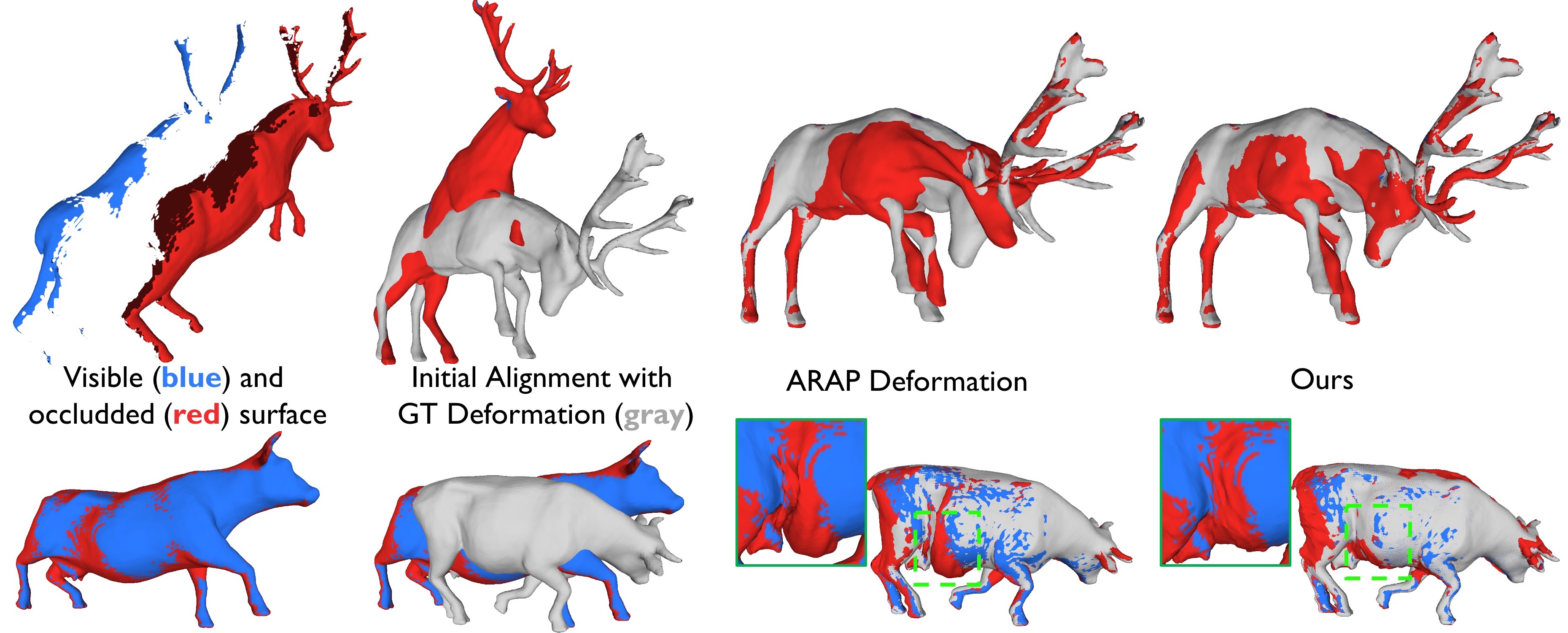}
\caption{
Surface deformation for the ``Deer'' and ``Dairy Cow'' sequence.
The complete shape and the motion of the visible surface (\blue{\textbf{blue}}) are given, and the goal is to estimate the deformation of the hidden surface (\red{\textbf{red}}). 
The gray mesh shows the ground truth deformation (\textcolor{mygray}{\textbf{gray}}), which is not available/used for registration.
ARAP leads to severe distortion on the neck and head of the deer; 
The dairy cow's stomach is undergoing a contraction movement. ARAP can not evenly distribute such deformation, leading to unnatural surface folding at the stomach. 
Our method yields natural deformations for both sequences. Note that our method is trained only on humanoid motions.
}
\label{fig:cow_deform}
\end{figure*}

\begin{table*}[!t]
\small
\centering

\renewcommand{\arraystretch}{1} % Default value: 1
\setlength{\tabcolsep}{6pt} % Default value: 6pt

\begin{tabular}{|l|c|c|c|c|c|c|c|}
\hline
Methods & 
    \makecell{ Humanoids \\ (Samba Dance) } & 
    \makecell{Dairy Cow \\ (Attack)}  & 
    \makecell{Moose Bull \\ (Walk)} &
    \makecell{ Fox\\ (Jump) }& 
    \makecell{Dear Stag\\(Attack)} & 
    \makecell{Panthera Onca \\ (Run) } & Avg.\\
\hline
Rigid Fitting                   				& 15.30        & 17.67         & 2.98      & 18.78   &16.96& 22.61   & 94.23     \\
\hline  
ARAP Deformation~\cite{arap} & 3.24          & 2.73          & 1.27         &  5.71    &4.99     & 13.78  & 31.72 \\
\hline
Motion Complete (Ours)          	& 2.32          &  2.90         & 1.34        &  4.88     &4.08    &8.56& 24.08\\       
\hline
Motion Complete + PP (Ours)    & \textbf{1.81}    & \textbf{2.24}        & \textbf{1.21}        & \textbf{4.20}    &\textbf{2.05}  &  \textbf{7.18}  & \textbf{18.69}  \\       
\hline
\end{tabular}
\vspace{0.1cm}
\caption{ 
Quantitative evaluation for the motion estimation results of the unobserved surface. 
The Metric is 3D End-Point-Error (EPE)  in centimeter.
Note that our method is trained only on humanoid motions. }
\label{tab:end_point_error_inv}
\end{table*}

\begin{table*}[]
\centering
\small
\renewcommand{\arraystretch}{1} % Default value: 1
\setlength{\tabcolsep}{12pt} % Default value: 6pt

\begin{tabular}{|l|c|c|c|c|c|c|} \hline                    
&         \multicolumn{3}{c|}{DeformingThings4D} & \multicolumn{3}{c|}{DeepDeform Dataset~\cite{deepdeform}} \\ \cline{2-7}
Method    & EPE$\downarrow$      & ACC(5)$\uparrow$     & ACC (10)$\uparrow$    & EPE$\downarrow$   & ACC(5)$\uparrow$  & ACC(10)$\uparrow$  \\ \hline 
Ours (\textit{w/o shape completion})     &3.82     & 79.02\%      & 90.55\%             & 13.75      &      26.89\%       &  63.42\%         \\   \hline 
Ours (\textit{w/ shape completion})              &\textbf{3.56}     & \textbf{85.02}\%      & \textbf{91.59}\% &   \textbf{13.15}  & \textbf{28.57\%}             & \textbf{63.66}\%           \\ \hline \end{tabular}
\caption{
Scene Flow estimation results on our DeformingThings4D dataset and DeepDeform~\cite{deepdeform} dataset.
All scores are reported only for the visible surface points.
Metrics are end-point-error (EPE) in centimeter, and accuracy ( \textless 5$cm$ or 5\%, 10$cm$ or 10\%).
}

\label{tab:end_point_error_vis}
\end{table*}

\begin{table}[h]
\small
\centering
\renewcommand{\arraystretch}{1} % Default value: 1
\setlength{\tabcolsep}{6pt} % Default value: 6pt

\begin{tabular}{|c|c|c|c|c|}
\hline
Method      			  & CD$\downarrow$ & IoU $\uparrow$    & SNC  $\uparrow$     & L1 $\downarrow$ \\ \hline
Ours (\textit{w/o motion})& 2.66 & 74.98\%  & 0.779   & 0.531 \\ \hline
Ours (\textit{w/ motion}) & \textbf{2.57} & \textbf{75.72}\%  & \textbf{0.812}   & \textbf{0.503} \\ \hline
\end{tabular}
\vspace{0.1cm}
\caption{ 
Surface prediction error on the test set of DeformingThings4D.
The metrics are Volumetric IoU (IoU), Chamfer Distance (CD) in centimeters, Surface Normal Consistency
(SNC), and $\ell_1$ score of SDF.}
\label{tab:geometry_score}
\end{table}

\subsection{Motion Prediction for the Hidden Surface}
This section evaluates the motion estimation of the hidden surface.
We conduct the following experiment: 
the complete mesh shape, a subset of mesh vertices that is visible from a given camera viewpoint, and the ground truth scene flow for the visible vertices are given, and the goal is to estimate the motion of the hidden vertices of the mesh.
We evaluate the following methods: 

\smallskip  
\noindent
\textbf{\textbullet Rigid Fitting.} This method assumes that the shape undergoes rigid motion.
It finds a single rigid transform in $SE(3)$ for the entire shape that best explains the surface motion.

\smallskip
\noindent
\textbf{\textbullet As-Rigid-As-Possible (ARAP) Deformation.} 
ARAP~\cite{arap} is widely used as a deformation prior in non-rigid reconstruction~\cite{dynamicfusion,volumedeform,zollhofer2014real}.
It assumes that locally, a point is transformed with a rigid transformation. 
Such rigid constraints are imposed upon nearby vertices that are connected by edges.
ARAP deformation finds for each mesh vertex a local fan-rotation $R\in SO(3)$ and a global translation vector $t\in \mathbb{R}^3$ that best explains the scene flow motion with the local rigidity constraints.

\smallskip
\noindent
\textbf{\textbullet Motion Complete (Ours).}
Given the complete shape and the partial motion on the visible surface, this method predicts the VMF for the complete shape and converts it to SFF to get the motion on mesh vertices' positions.
This method is trained only on humanoid motions and evaluates on an animal motion subset (we aim to confirm how the model generalizes across domains). 

\smallskip
\noindent
\textbf{\textbullet Motion Complete + Post Processing (PP)  (Ours).}
We found that the motion prediction of our Motion Complete model is sometimes noisy. 
We employ optimization-based post-processing to alleviate the noise: the predicted motion filed on the mesh surface is jointly optimized with ARAP prior that enforce that nearby vertices have similar motions.

\smallskip
Tab.~\ref{tab:end_point_error_inv} reports the motion estimation results for the occluded surface. The testing sequence includes one humanoid sequence and 6 animal sequences with different animations.
Note that our method is trained only on the humanoids dataset.
Among the baselines, rigid fitting yields significantly larger errors on most sequences, which indicate that the sequences undergo large non-rigid motion. 
Our Motion Complete overall achieves lower end-point-error than the ARAP on most sequences.
Motion Complete + PP further improves the numbers. 
Fig.~\ref{fig:cow_deform} shows the qualitative results of surface deformation for the ``Deer'' and ``Dairy Cow'' sequence. 
The deformed surfaces are achieved by using the estimated motion to warp the source model.
Our method yields more plausible deformation than ARAP deformation for the occluded surface. 
We conclude that the 3D sparse ConvNets with large receptive fields learn to capture the global deformation.

\subsection{Real-World Results}
Fig.~\ref{fig:demo_screenshot} shows that our method, trained only on our synthetic data, generalizes well to a real-world RGB-D input captured with an Azure Kinect camera. 

\subsection{Ablation of Shape and Motion Estimation}
This experiment examines how the two tasks, geometry completion and motion estimation, influence each other.
To get the scene flow of the visible surface, we re-train FlowNet3D~\cite{flownet3d} using our scene flow dataset. 
FlowNet3D predicts the SFF given a pair of point clouds with a subsampled size of 2048. 
We convert the sparse SFF to VMF using Eqn.~\ref{eqn:sff_2_vmf} as network input.
The voxel position in the VMF is consistent with the input projective TSDF.

As defined in Fig.~\ref{fig:network_architecture}, we alternatively remove the shape completion head or the motion estimation head to examine the synergy of the two tasks.
Tab.~\ref{tab:end_point_error_vis} reports the motion prediction results for the visible surface.
Though only evaluating on the visible surface, the model trained with the added supervision of the geometry completion task show improvement over the model trained only on motion prediction.
This demonstrates that complete the missing shape is beneficial for non-rigid motion estimation.
Tab.~\ref{tab:geometry_score} reports the geometry completion results in our synthetic DeformingThings4D dataset. 
The whole model shows improvement over the model that is trained for geometry completion only.
This result validates the idea that in a dynamic scene it is beneficial to understand the motion in order to achieve better geometric completion.

\begin{figure}[!ht]
\centering
\includegraphics[width= 1.02\linewidth ]{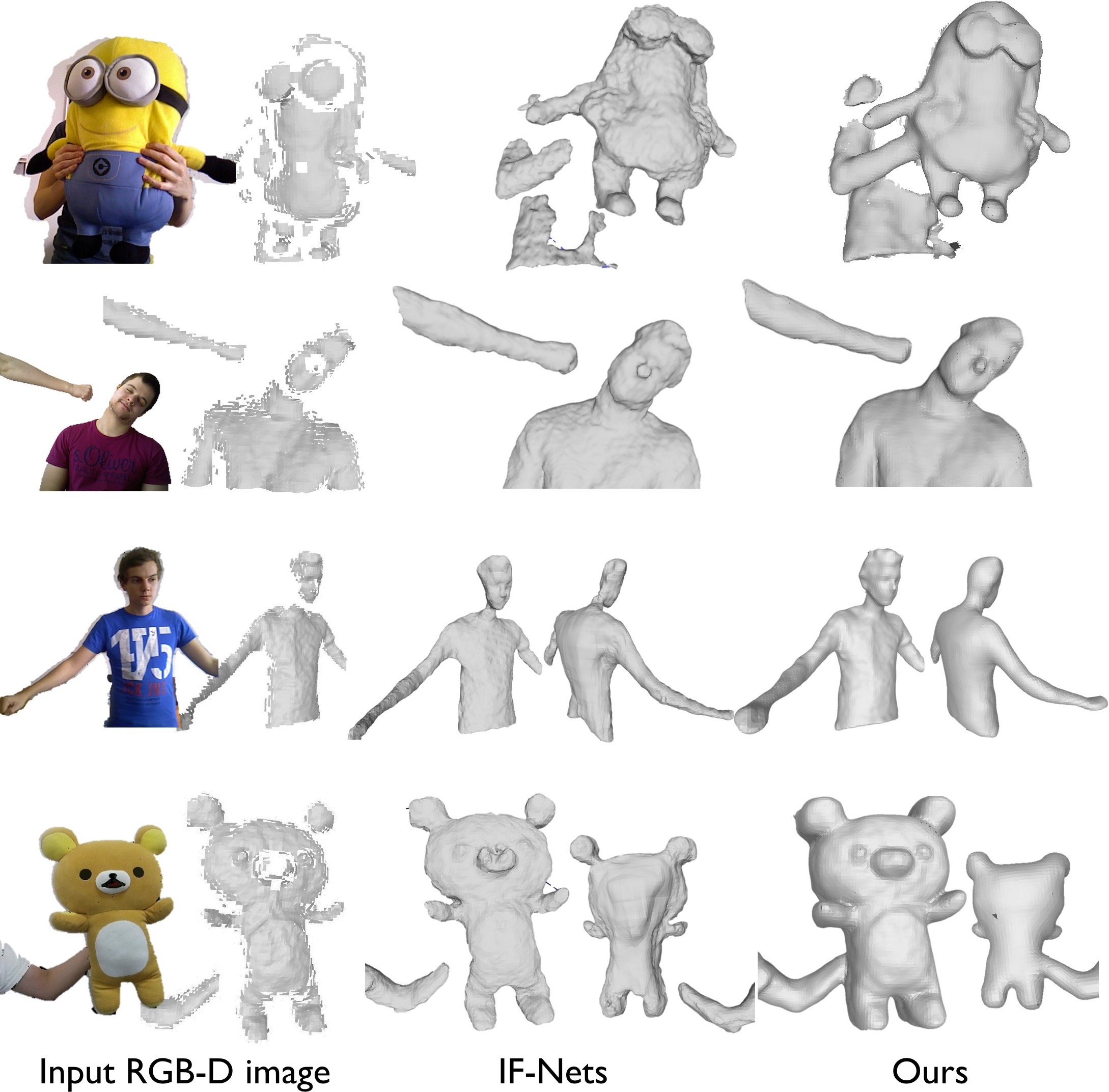}
\caption{Shape completion results on real-world RGB-D images.
The top 3 rows are images from VolumeDeform~\cite{volumedeform}, and the last row is from Li et al.~\cite{Learning2Optimize}.
}
\label{fig:sgnn_real}
\end{figure}

 \begin{figure}[!h]
\centering
\includegraphics[width= 1.12\linewidth ]{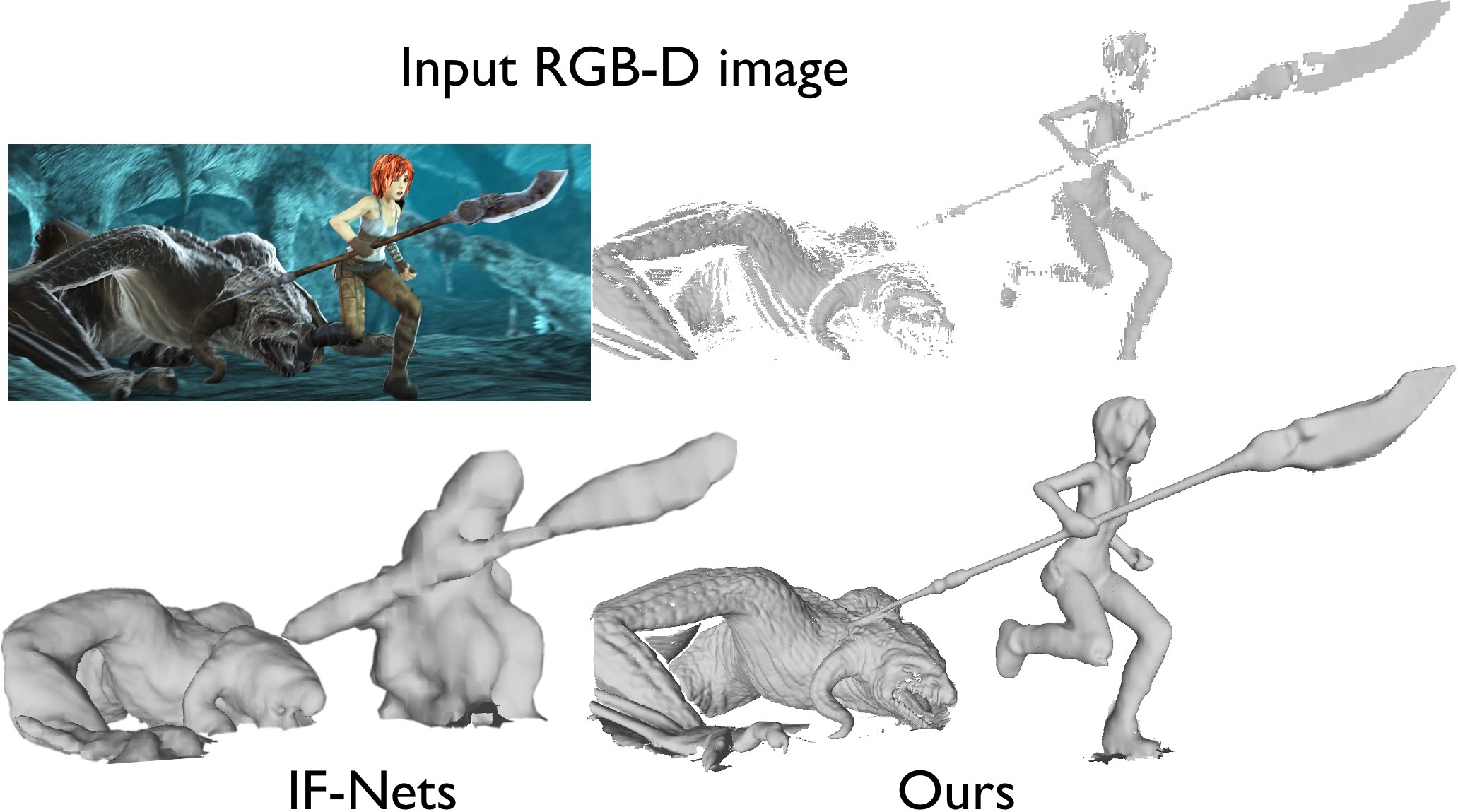}
\caption{
Shape completion on  a large scene.
The image is from MPI Sintel~\cite{Sintel} dataset. The maximum depth is set to 10 meters in this sintel scene.
With fixed volume size ($256^3$), IF-Net~\cite{chibane20ifnet} loses the ability to model details for large scenes.  
Our fully-convolutional approach better captures all levels of detail.
}
\label{fig:sgnn_sintel}
\end{figure}

\subsection{Shape Completion Results}
We show qualitative shape completion results of our approach.
IF-Nets~\cite{chibane20ifnet} is a state-of-the-art method that performs single depth image reconstruction from point clouds.
At the core of IF-Nets is an implicit function that maps a 3D coordinate to an occupancy score using a multi-layer perceptron.
We train both methods on the humanoids dataset and evaluate the completion performance on unseen sequences.
Fig.~\ref{fig:sgnn_real} shows shape completion from real-world RGB-D images. 
Our fully-convolutional approach shows more complete, sharper results than the implicit IF-Net.
Tab.~\ref{tab:volumedeform_number} shows quantitative evaluation on VolumeDeform~\cite{volumedeform} sequences.
In particular, for large scenes, our approach effectively captures both global and local structures, as shown in Fig.~\ref{fig:sgnn_sintel}.

\begin{table}[!h]
\centering
\small
\begin{tabular}{|l|c|c|c|}
\hline
Methods  & IF-Nets & Ours (w/o motion) & Ours  \\
\hline
P2P (cm)$\downarrow$ & 2.231   & 1.983       & \textbf{1.876}                 \\
\hline
SNC $\uparrow$  & 0.757   & 0.899             & \textbf{0.908}                 \\
\hline
time (s) $\downarrow$ & 14.26 & \textbf{3.19} & 3.45\\ 
\hline
memory (MB) $\downarrow$ & 19,437& \textbf{1,103} & 1,379\\
\hline
\end{tabular}
\caption{
Quantitative results on VolumeDeform examples.
The surface ground-truth is provided by VolumeDeform. The metrics are the point-to-plane (P2P) distance, and surface normal consistency (SNC).
We also report the average time and memory required for the inference on a Tesla V100-SXM2-32GB GPU. 
}
\label{tab:volumedeform_number}
\end{table}

\paragraph{Limitations}
Our approach maintains several limitations:
1) estimating the uncertainty of hidden motion is necessary but not handled by our approach. A probabilistic approach (e.g.,  Huang et al.~\cite{huang2020di}) would be  promising for modeling motion uncertainty.
2) our method does not predict surface colors. 
The differentiable volumetric rendering approach of~\cite{dai2020spsg} is a potential solution to learn colored deformable objects.
3) DeformingThings4D largely contains articulated objects such as humans and animals. We are planning to expand the dataset with examples of loose clothing or plants that are deformed by external forces.

\section{Conclusion}
In this work, we present the first method that jointly estimates the invisible shape and deformation from partial depth frame observation. 
We show that shape completion and motion estimation are mutually complementary tasks, with joint learning benefiting each.
Our newly proposed animation dataset allows for cross-domain generalization for both motion and shape. We believe that our method and new dataset open a new research avenue on generic non-rigid 4D reconstruction.

\section{Acknowledgements}
This work was conducted during Yang Li's internship at Tokyo Research Center, Huawei. 
Matthias Nie{\ss}ner is supported by a TUM-IAS Rudolf M{\"o}{\ss}bauer Fellowship and the ERC Starting Grant Scan2CAD (804724). 
We thank Angela Dai for the voice-over of the video and also thank for setting up the DeformingThings4D dataset.

{\small
\bibliographystyle{ieee_fullname}
\bibliography{yang_bib_2021}

\begin{thebibliography}{10}\itemsep=-1pt

\bibitem{NICP-1}
Brian Amberg, Sami Romdhani, and Thomas Vetter.
\newblock Optimal step nonrigid icp algorithms for surface registration.
\newblock In {\em Proceedings of the IEEE International Conference on Computer
  Vision (ICCV)}, pages 1--8. IEEE, 2007.

\bibitem{anguelov2005scape}
Dragomir Anguelov, Praveen Srinivasan, Daphne Koller, Sebastian Thrun, Jim
  Rodgers, and James Davis.
\newblock Scape: shape completion and animation of people.
\newblock In {\em ACM SIGGRAPH 2005 Papers}, pages 408--416. 2005.

\bibitem{faust}
Federica Bogo, Javier Romero, Matthew Loper, and Michael~J Black.
\newblock Faust: Dataset and evaluation for 3d mesh registration.
\newblock In {\em Proceedings of the IEEE Conference on Computer Vision and
  Pattern Recognition (CVPR)}, pages 3794--3801, 2014.

\bibitem{bovzivc2020neural}
Alja{\v{z}} Bo{\v{z}}i{\v{c}}, Pablo Palafox, Michael Zollh{\"o}fer, Angela
  Dai, Justus Thies, and Matthias Nie{\ss}ner.
\newblock Neural non-rigid tracking.
\newblock {\em arXiv preprint arXiv:2006.13240}, 2020.

\bibitem{deepdeform}
Alja{\v{z}} Bo{\v{z}}i{\v{c}}, Michael Zollh{\"o}fer, Christian Theobalt, and
  Matthias Nie{\ss}ner.
\newblock Deepdeform: Learning non-rigid rgb-d reconstruction with
  semi-supervised data.
\newblock In {\em Proceedings of the IEEE Conference on Computer Vision and
  Pattern Recognition (CVPR)}, pages 7002--7012, 2020.

\bibitem{Sintel}
Daniel~J Butler, Jonas Wulff, Garrett~B Stanley, and Michael~J Black.
\newblock A naturalistic open source movie for optical flow evaluation.
\newblock In {\em Proceedings of the European Conference on Computer Vision
  (ECCV)}, pages 611--625, 2012.

\bibitem{chibane20ifnet}
Julian Chibane, Thiemo Alldieck, and Gerard Pons-Moll.
\newblock Implicit functions in feature space for 3d shape reconstruction and
  completion.
\newblock In {\em Proceedings of the IEEE Conference on Computer Vision and
  Pattern Recognition (CVPR)}, pages 6970--6981, 2020.

\bibitem{choy2019minkowski}
Christopher Choy, JunYoung Gwak, and Silvio Savarese.
\newblock 4d spatio-temporal convnets: Minkowski convolutional neural networks.
\newblock In {\em Proceedings of the IEEE Conference on Computer Vision and
  Pattern Recognition (CVPR)}, pages 3075--3084, 2019.

\bibitem{volumetric_fusion}
Brian Curless and Marc Levoy.
\newblock A volumetric method for building complex models from range images.
\newblock In {\em Proceedings of the 23rd annual conference on Computer
  graphics and interactive techniques}, pages 303--312, 1996.

\bibitem{sgnn_cvpr2020}
Angela Dai, Christian Diller, and Matthias Nie{\ss}ner.
\newblock Sg-nn: Sparse generative neural networks for self-supervised scene
  completion of rgb-d scans.
\newblock In {\em Proceedings of the IEEE Conference on Computer Vision and
  Pattern Recognition (CVPR)}, pages 849--858, 2020.

\bibitem{scancomplete}
Angela Dai, Daniel Ritchie, Martin Bokeloh, Scott Reed, J{\"u}rgen Sturm, and
  Matthias Nie{\ss}ner.
\newblock Scancomplete: Large-scale scene completion and semantic segmentation
  for 3d scans.
\newblock In {\em Proceedings of the IEEE Conference on Computer Vision and
  Pattern Recognition (CVPR)}, pages 4578--4587, 2018.

\bibitem{3d_epn}
Angela Dai, Charles Ruizhongtai~Qi, and Matthias Nie{\ss}ner.
\newblock Shape completion using 3d-encoder-predictor cnns and shape synthesis.
\newblock In {\em Proceedings of the IEEE Conference on Computer Vision and
  Pattern Recognition (CVPR)}, pages 5868--5877, 2017.

\bibitem{dai2020spsg}
Angela Dai, Yawar Siddiqui, Justus Thies, Julien Valentin, and Matthias
  Nie{\ss}ner.
\newblock Spsg: Self-supervised photometric scene generation from rgb-d scans.
\newblock {\em arXiv preprint arXiv:2006.14660}, 2020.

\bibitem{de2008performance}
Edilson De~Aguiar, Carsten Stoll, Christian Theobalt, Naveed Ahmed, Hans-Peter
  Seidel, and Sebastian Thrun.
\newblock Performance capture from sparse multi-view video.
\newblock In {\em ACM SIGGRAPH 2008 papers}, pages 1--10, 2008.

\bibitem{flownet}
Alexey Dosovitskiy, Philipp Fischer, Eddy Ilg, Philip Hausser, Caner Hazirbas,
  Vladimir Golkov, Patrick Van Der~Smagt, Daniel Cremers, and Thomas Brox.
\newblock Flownet: Learning optical flow with convolutional networks.
\newblock In {\em Proceedings of the IEEE International Conference on Computer
  Vision (ICCV)}, pages 2758--2766, 2015.

\bibitem{motion2fusion}
Mingsong Dou, Philip Davidson, Sean~Ryan Fanello, Sameh Khamis, Adarsh Kowdle,
  Christoph Rhemann, Vladimir Tankovich, and Shahram Izadi.
\newblock Motion2fusion: Real-time volumetric performance capture.
\newblock {\em ACM Transactions on Graphics (TOG)}, 36(6):1--16, 2017.

\bibitem{dou2016fusion4d}
Mingsong Dou, Sameh Khamis, Yury Degtyarev, Philip Davidson, Sean~Ryan Fanello,
  Adarsh Kowdle, Sergio~Orts Escolano, Christoph Rhemann, David Kim, Jonathan
  Taylor, et~al.
\newblock Fusion4d: Real-time performance capture of challenging scenes.
\newblock {\em ACM Transactions on Graphics (TOG)}, 35(4):114, 2016.

\bibitem{surfelwarp}
Wei Gao and Russ Tedrake.
\newblock Surfelwarp: Efficient non-volumetric single view dynamic
  reconstruction.
\newblock {\em arXiv preprint arXiv:1904.13073}, 2019.

\bibitem{kitti}
Andreas Geiger, Philip Lenz, and Raquel Urtasun.
\newblock Are we ready for autonomous driving? the kitti vision benchmark
  suite.
\newblock In {\em Proceedings of the IEEE Conference on Computer Vision and
  Pattern Recognition (CVPR)}, pages 3354--3361, 2012.

\bibitem{sparseconv_iccv}
Benjamin Graham, Martin Engelcke, and Laurens Van Der~Maaten.
\newblock 3d semantic segmentation with submanifold sparse convolutional
  networks.
\newblock In {\em Proceedings of the IEEE Conference on Computer Vision and
  Pattern Recognition (CVPR)}, pages 9224--9232, 2018.

\bibitem{sparseconv_arxiv}
Benjamin Graham and Laurens van~der Maaten.
\newblock Submanifold sparse convolutional networks.
\newblock {\em arXiv preprint arXiv:1706.01307}, 2017.

\bibitem{HPLFlownet}
Xiuye Gu, Yijie Wang, Chongruo Wu, Yong~Jae Lee, and Panqu Wang.
\newblock Hplflownet: Hierarchical permutohedral lattice flownet for scene flow
  estimation on large-scale point clouds.
\newblock In {\em Proceedings of the IEEE Conference on Computer Vision and
  Pattern Recognition (CVPR)}, pages 3254--3263, 2019.

\bibitem{guo2015robust}
Kaiwen Guo, Feng Xu, Yangang Wang, Yebin Liu, and Qionghai Dai.
\newblock Robust non-rigid motion tracking and surface reconstruction using l0
  regularization.
\newblock In {\em Proceedings of the IEEE International Conference on Computer
  Vision (ICCV)}, pages 3083--3091, 2015.

\bibitem{WholeIsGreaterThanParts}
Oshri Halimi, Ido Imanuel, Or Litany, Giovanni Trappolini, Emanuele Rodol{\`a},
  Leonidas Guibas, and Ron Kimmel.
\newblock The whole is greater than the sum of its nonrigid parts.
\newblock {\em arXiv preprint arXiv:2001.09650}, 2020.

\bibitem{huang2020di}
Jiahui Huang, Shi-Sheng Huang, Haoxuan Song, and Shi-Min Hu.
\newblock Di-fusion: Online implicit 3d reconstruction with deep priors.
\newblock {\em arXiv preprint arXiv:2012.05551}, 2020.

\bibitem{volumedeform}
Matthias Innmann, Michael Zollh{\"o}fer, Matthias Nie{\ss}ner, Christian
  Theobalt, and Marc Stamminger.
\newblock Volumedeform: Real-time volumetric non-rigid reconstruction.
\newblock In {\em Proceedings of the European Conference on Computer Vision
  (ECCV)}, pages 362--379, 2016.

\bibitem{chiyu2020localimplicit}
Chiyu Jiang, Avneesh Sud, Ameesh Makadia, Jingwei Huang, Matthias Nie{\ss}ner,
  and Thomas Funkhouser.
\newblock Local implicit grid representations for 3d scenes.
\newblock In {\em Proceedings of the IEEE Conference on Computer Vision and
  Pattern Recognition (CVPR)}, pages 6001--6010, 2020.

\bibitem{PoissonSurfaceReconst}
Michael Kazhdan, Matthew Bolitho, and Hugues Hoppe.
\newblock Poisson surface reconstruction.
\newblock In {\em Proceedings of the fourth Eurographics symposium on Geometry
  processing}, volume~7, 2006.

\bibitem{li2008regist}
Hao Li, Robert~W Sumner, and Mark Pauly.
\newblock Global correspondence optimization for non-rigid registration of
  depth scans.
\newblock In {\em Computer graphics forum}, volume~27, pages 1421--1430. Wiley
  Online Library, 2008.

\bibitem{Learning2Optimize}
Yang Li, Aljaz Bozic, Tianwei Zhang, Yanli Ji, Tatsuya Harada, and Matthias
  Nie{\ss}ner.
\newblock Learning to optimize non-rigid tracking.
\newblock In {\em Proceedings of the IEEE Conference on Computer Vision and
  Pattern Recognition (CVPR)}, pages 4910--4918, 2020.

\bibitem{PiFusion}
Zhe Li, Tao Yu, Chuanyu Pan, Zerong Zheng, and Yebin Liu.
\newblock Robust 3d self-portraits in seconds.
\newblock In {\em Proceedings of the IEEE Conference on Computer Vision and
  Pattern Recognition (CVPR)}, pages 1344--1353, 2020.

\bibitem{flownet3d}
Xingyu Liu, Charles~R Qi, and Leonidas~J Guibas.
\newblock Flownet3d: Learning scene flow in 3d point clouds.
\newblock In {\em Proceedings of the IEEE Conference on Computer Vision and
  Pattern Recognition (CVPR)}, pages 529--537, 2019.

\bibitem{liu2019meteornet}
Xingyu Liu, Mengyuan Yan, and Jeannette Bohg.
\newblock Meteornet: Deep learning on dynamic 3d point cloud sequences.
\newblock In {\em Proceedings of the IEEE International Conference on Computer
  Vision (ICCV)}, pages 9246--9255, 2019.

\bibitem{lv2018LearningRigidity}
Zhaoyang Lv, Kihwan Kim, Alejandro Troccoli, Deqing Sun, James~M Rehg, and Jan
  Kautz.
\newblock Learning rigidity in dynamic scenes with a moving camera for 3d
  motion field estimation.
\newblock In {\em Proceedings of the European Conference on Computer Vision
  (ECCV)}, pages 468--484, 2018.

\bibitem{AMASS_ICCV2019}
Naureen Mahmood, Nima Ghorbani, Nikolaus~F. Troje, Gerard Pons-Moll, and
  Michael~J. Black.
\newblock Amass: Archive of motion capture as surface shapes.
\newblock In {\em Proceedings of the IEEE International Conference on Computer
  Vision (ICCV)}, Oct 2019.

\bibitem{Monka}
Nikolaus Mayer, Eddy Ilg, Philip Hausser, Philipp Fischer, Daniel Cremers,
  Alexey Dosovitskiy, and Thomas Brox.
\newblock A large dataset to train convolutional networks for disparity,
  optical flow, and scene flow estimation.
\newblock In {\em Proceedings of the IEEE Conference on Computer Vision and
  Pattern Recognition (CVPR)}, pages 4040--4048, 2016.

\bibitem{occupancy_network}
Lars Mescheder, Michael Oechsle, Michael Niemeyer, Sebastian Nowozin, and
  Andreas Geiger.
\newblock Occupancy networks: Learning 3d reconstruction in function space.
\newblock In {\em Proceedings of the IEEE Conference on Computer Vision and
  Pattern Recognition (CVPR)}, pages 4460--4470, 2019.

\bibitem{dynamicfusion}
Richard~A Newcombe, Dieter Fox, and Steven~M Seitz.
\newblock Dynamicfusion: Reconstruction and tracking of non-rigid scenes in
  real-time.
\newblock In {\em Proceedings of the IEEE Conference on Computer Vision and
  Pattern Recognition (CVPR)}, pages 343--352, 2015.

\bibitem{occupancy_flow}
Michael Niemeyer, Lars Mescheder, Michael Oechsle, and Andreas Geiger.
\newblock Occupancy flow: 4d reconstruction by learning particle dynamics.
\newblock In {\em Proceedings of the IEEE International Conference on Computer
  Vision (ICCV)}, pages 5379--5389, 2019.

\bibitem{deepsdf}
Jeong~Joon Park, Peter Florence, Julian Straub, Richard Newcombe, and Steven
  Lovegrove.
\newblock Deepsdf: Learning continuous signed distance functions for shape
  representation.
\newblock In {\em Proceedings of the IEEE Conference on Computer Vision and
  Pattern Recognition (CVPR)}, pages 165--174, 2019.

\bibitem{NICP-2}
Mark Pauly, Niloy~J Mitra, Joachim Giesen, Markus~H Gross, and Leonidas~J
  Guibas.
\newblock Example-based 3d scan completion.
\newblock In {\em Symposium on Geometry Processing}, pages 23--32, 2005.

\bibitem{convolutional_occnet}
Songyou Peng, Michael Niemeyer, Lars Mescheder, Marc Pollefeys, and Andreas
  Geiger.
\newblock Convolutional occupancy networks.
\newblock {\em arXiv preprint arXiv:2003.04618}, 2020.

\bibitem{pointnet++}
Charles~Ruizhongtai Qi, Li Yi, Hao Su, and Leonidas~J Guibas.
\newblock Pointnet++: Deep hierarchical feature learning on point sets in a
  metric space.
\newblock In {\em Advances in neural information processing systems}, pages
  5099--5108, 2017.

\bibitem{playing_for_benchmarks}
Stephan~R Richter, Zeeshan Hayder, and Vladlen Koltun.
\newblock Playing for benchmarks.
\newblock In {\em Proceedings of the IEEE International Conference on Computer
  Vision (ICCV)}, pages 2213--2222, 2017.

\bibitem{killingfusion}
Miroslava Slavcheva, Maximilian Baust, Daniel Cremers, and Slobodan Ilic.
\newblock Killingfusion: Non-rigid 3d reconstruction without correspondences.
\newblock In {\em Proceedings of the IEEE Conference on Computer Vision and
  Pattern Recognition (CVPR)}, pages 1386--1395, 2017.

\bibitem{SSCNet_shuran}
Shuran Song, Fisher Yu, Andy Zeng, Angel~X Chang, Manolis Savva, and Thomas
  Funkhouser.
\newblock Semantic scene completion from a single depth image.
\newblock In {\em Proceedings of the IEEE Conference on Computer Vision and
  Pattern Recognition (CVPR)}, pages 1746--1754, 2017.

\bibitem{arap}
Olga Sorkine and Marc Alexa.
\newblock As-rigid-as-possible surface modeling.
\newblock In {\em Symposium on Geometry Processing}, volume~4, pages 109--116,
  2007.

\bibitem{random_orthogonal}
Gilbert~W Stewart.
\newblock The efficient generation of random orthogonal matrices with an
  application to condition estimators.
\newblock {\em SIAM Journal on Numerical Analysis}, 17(3):403--409, 1980.

\bibitem{embededdeformation}
Robert~W Sumner, Johannes Schmid, and Mark Pauly.
\newblock Embedded deformation for shape manipulation.
\newblock {\em ACM Transactions on Graphics (TOG)}, 26(3):80, 2007.

\bibitem{pwcnet}
Deqing Sun, Xiaodong Yang, Ming-Yu Liu, and Jan Kautz.
\newblock Pwc-net: Cnns for optical flow using pyramid, warping, and cost
  volume.
\newblock In {\em Proceedings of the IEEE Conference on Computer Vision and
  Pattern Recognition (CVPR)}, pages 8934--8943, 2018.

\bibitem{RAFT_eccv2020}
Zachary Teed and Jia Deng.
\newblock Raft: Recurrent all-pairs field transforms for optical flow.
\newblock {\em arXiv preprint arXiv:2003.12039}, 2020.

\bibitem{vlasic2008articulated}
Daniel Vlasic, Ilya Baran, Wojciech Matusik, and Jovan Popovi{\'c}.
\newblock Articulated mesh animation from multi-view silhouettes.
\newblock In {\em ACM SIGGRAPH 2008 papers}, pages 1--9, 2008.

\bibitem{global_patch_collider}
Shenlong Wang, Sean Ryan~Fanello, Christoph Rhemann, Shahram Izadi, and
  Pushmeet Kohli.
\newblock The global patch collider.
\newblock In {\em Proceedings of the IEEE Conference on Computer Vision and
  Pattern Recognition (CVPR)}, pages 127--135, 2016.

\bibitem{flownet3d++}
Zirui Wang, Shuda Li, Henry Howard-Jenkins, Victor Prisacariu, and Min Chen.
\newblock Flownet3d++: Geometric losses for deep scene flow estimation.
\newblock In {\em The IEEE Winter Conference on Applications of Computer
  Vision}, pages 91--98, 2020.

\bibitem{wu2019pointpwc}
Wenxuan Wu, Zhiyuan Wang, Zhuwen Li, Wei Liu, and Li Fuxin.
\newblock Pointpwc-net: A coarse-to-fine network for supervised and
  self-supervised scene flow estimation on 3d point clouds.
\newblock {\em arXiv preprint arXiv:1911.12408}, 2019.

\bibitem{ye2012performance}
Genzhi Ye, Yebin Liu, Nils Hasler, Xiangyang Ji, Qionghai Dai, and Christian
  Theobalt.
\newblock Performance capture of interacting characters with handheld kinects.
\newblock In {\em Proceedings of the European Conference on Computer Vision
  (ECCV)}, pages 828--841, 2012.

\bibitem{zheng2013beyond}
Bo Zheng, Yibiao Zhao, Joey~C Yu, Katsushi Ikeuchi, and Song-Chun Zhu.
\newblock Beyond point clouds: Scene understanding by reasoning geometry and
  physics.
\newblock In {\em Proceedings of the IEEE Conference on Computer Vision and
  Pattern Recognition (CVPR)}, pages 3127--3134, 2013.

\bibitem{Zheng2019DeepHuman}
Zerong Zheng, Tao Yu, Yixuan Wei, Qionghai Dai, and Yebin Liu.
\newblock Deephuman: 3d human reconstruction from a single image.
\newblock In {\em Proceedings of the IEEE International Conference on Computer
  Vision (ICCV)}, October 2019.

\bibitem{zollhofer2014real}
Michael Zollh{\"o}fer, Matthias Nie{\ss}ner, Shahram Izadi, Christoph Rehmann,
  Christopher Zach, Matthew Fisher, Chenglei Wu, Andrew Fitzgibbon, Charles
  Loop, Christian Theobalt, et~al.
\newblock Real-time non-rigid reconstruction using an rgb-d camera.
\newblock {\em ACM Transactions on Graphics (ToG)}, 33(4):156, 2014.

\bibitem{3D_menagerie}
Silvia Zuffi, Angjoo Kanazawa, David~W Jacobs, and Michael~J Black.
\newblock 3d menagerie: Modeling the 3d shape and pose of animals.
\newblock In {\em Proceedings of the IEEE Conference on Computer Vision and
  Pattern Recognition (CVPR)}, pages 6365--6373, 2017.

\end{thebibliography}
}

\end{document}